\documentclass[runningheads]{llncs}
\usepackage{microtype}
\usepackage[pdftex]{graphicx}
\usepackage{epstopdf}
\usepackage[hyphens]{url}

\usepackage{subfigure}
\usepackage{algorithm}
\usepackage{algorithmic}
\usepackage{booktabs} 

\usepackage{amsfonts}
\usepackage{amsmath}
\DeclareMathOperator*{\argmax}{argmax}
\usepackage[misc, geometry]{ifsym}
\begin{document}
\title{A Framework for Deep Constrained Clustering - Algorithms and Advances}
%
%
\author{Hongjing Zhang \inst{1}(\Letter)\and
Sugato Basu\inst{2} \and
Ian Davidson\inst{1}}
\authorrunning{H.J. Zhang, S. Basu and I. Davidson}
%
\institute{
Department of Computer Science, University of California, Davis, CA 95616, USA \\ 
\email {hjzzhang@ucdavis.edu}, \email{davidson@cs.ucdavis.edu} \and
Google Research,
Mountain View, CA 94043, USA \\
\email{sugato@google.com}\\
}

\maketitle              
\begin{abstract}
The area of constrained clustering has been extensively explored by researchers and used by practitioners. Constrained clustering formulations exist for popular algorithms such as k-means, mixture models, and spectral clustering but have several limitations. A fundamental strength of deep learning is its flexibility, and here we explore a deep learning framework for constrained clustering and in particular explore how it can extend the field of constrained clustering. We show that our framework can not only handle standard together/apart constraints (without the well documented negative effects reported earlier) generated from labeled side information but more complex constraints generated from new types of side information such as continuous values and high-level domain knowledge. (Source code available at: \url{http://github.com/blueocean92/deep_constrained_clustering})

\end{abstract}

\section{Introduction}
\label{introduction}
Constrained clustering has a long history in machine learning with many standard algorithms being adapted to be constrained \cite{basu2008constrained} including EM \cite{basu2004probabilistic}, K-Means \cite{wagstaff2001constrained} and spectral methods \cite{wang2010flexible}. The addition of constraints generated from ground truth labels allows a semi-supervised setting to increase accuracy \cite{wagstaff2001constrained} when measured against the ground truth labeling.

However, there are several limitations in these methods and one purpose of this paper is to explore how deep learning can make advances to the field beyond what other methods have. In particular, we find that existing non-deep formulations of constrained clustering have the following limitations:
\begin{itemize}
\item \emph{Limited Constraints and Side Information}. Constraints are limited to simple together/apart constraints typically generated from labels. In some domains, experts may more naturally give guidance at the cluster level or generate constraints from continuous side-information.
\item \emph{Negative Effect of Constraints.} For some algorithms though  constraints improve performance when \emph{averaged} over many constraint sets, \emph{individual} constraint sets produce results worse than using no constraints \cite{davidson2006measuring}. As practitioners typically have one constraint set their use can be ``hit or miss".
\item \emph{Intractability and Scalability Issues.} Iterative algorithms that directly solve for clustering assignments run into problems of intractability \cite{davidson2007intractability}. Relaxed formulations (i.e. spectral methods \cite{lu2008constrained,wang2010flexible}) require solving a full rank eigendecomposition problem which takes $O(n^3)$. 
\item \emph{Assumption of Good Features}. A core requirement is that good features or similarity function for complex data is already created. 

\end{itemize}

Since deep learning is naturally scalable and able to find useful representations we focus on the first and second challenges but experimentally explore the third and fourth. Though deep clustering with constraints has many potential benefits to overcome these limitations it is not without its challenges. Our major contributions in this paper are summarized as follows:
\begin{itemize}
\item We propose a deep constrained clustering formulation that cannot only encode standard together/apart constraints but new triplet constraints (which can be generated from continuous side information), instance difficulty constraints, and cluster level balancing constraints (see section \ref{sec:methods}). 
    \item Deep constrained clustering overcomes a long term issue we reported in PKDD earlier \cite{davidson2006measuring} with constrained clustering of profound practical implications: overcoming the negative effects of individual constraint sets. 
    \item We show how the benefits of deep learning such as scalability and end-to-end learning translate to our deep constrained clustering formulation. We achieve better clustering results than traditional constrained clustering methods (with features generated from an auto-encoder) on challenging datasets (see Table \ref{pairwise_neg}).
\end{itemize}

Our paper is organized as follows. First, we introduce the related work in section \ref{sec:related work}. We then propose four forms of constraints in section \ref{sec:methods} and introduce how to train the clustering network with these constraints in section \ref{sec:training}. Then we compare our approach to previous baselines and demonstrate the effectiveness of new types of constraints in section \ref{sec:experiments}. Finally, we discuss future work and conclude in section \ref{sec:conclusion}.

\section{Related Work}
\label{sec:related work}
\textbf{Constrained Clustering.} Constrained clustering is an important area and there is a large body of work that shows how \emph{side information} can improve the clustering performance \cite{wagstaff2000clustering,wagstaff2001constrained,xing2003distance,bilenko2004integrating,wang2010flexible}. Here the side information is typically labeled data which is used to generate \emph{pairwise} together/apart constraints used to partially reveal the ground truth clustering to help the clustering algorithm. Such constraints are easy to encode in matrices and enforce in procedural algorithms though not with its challenges. In particular, we showed \cite{davidson2006measuring} performance improves with larger constraint sets when \textbf{averaged} over many constraint sets generated from the ground truth labeling. However, for a significant fraction (just not the majority) of these constraint sets performance is \emph{worse} than using no constraint set. We recreated some of these results in Table \ref{pairwise_neg}.

Moreover, side information can exist in different forms beyond labels (i.e. continuous data), and domain experts can provide guidance beyond pairwise constraints. Some work in the supervised classification setting \cite{joachims2002optimizing,schultz2004learning,schroff2015facenet,gress2016probabilistic} seek alternatives such as relative/triplet guidance, but to our knowledge, such information has not been explored in the non-hierarchical clustering setting. Complex constraints for hierarchical clustering have been explored \cite{bade2008creating,chatziafratis2018hierarchical} but these are tightly limited to the hierarchical structure (i.e., $x$ must be joined with $y$ before $z$) and not directly translated to non-hierarchical (partitional) clustering.

\textbf{Deep Clustering.} Motivated by the success of deep neural networks in supervised learning, unsupervised deep learning approaches are now being explored \cite{xie2016unsupervised,jiang2016variational,yang2016towards,guo2017improved,shaham2018spectralnet}. There are approaches \cite{yang2016towards,shaham2018spectralnet} which learn an encoding that is suitable for a clustering objective first and then applied an external clustering method. Our work builds upon the most direct setting \cite{xie2016unsupervised,guo2017improved} which encodes one self-training objective and finds the clustering allocations for all instances within one neural network.

\textbf{Deep Clustering with Pairwise Constraints.}  Most recently, the semi-supervised clustering networks with pairwise constraints have been explored: \cite{hsu2015neural} uses pairwise constraints to enforce small divergence between similar pairs while increasing the divergence between dissimilar pairs assignment probability distributions. However, this approach did not leverage the unlabeled data, hence requires lot's of labeled data to achieve good results. Fogel et al. proposed an unsupervised clustering network \cite{fogel2018clustering} by self-generating pairwise constraints from mutual KNN graph and extends it to semi-supervised clustering by using labeled connections queried from the human. However, this method cannot make out-of-sample predictions and requires user-defined parameters for generating constraints from mutual KNN graph. 

\section{Deep Constrained Clustering Framework}
\label{sec:methods}
Here we outline our proposed framework for deep constrained clustering. Our method of adding constraints to and training deep learning can be used for most deep clustering method (so long as the network has a $k$ unit output indicating the degree of cluster membership) and here we choose the popular deep embedded clustering method (DEC \cite{xie2016unsupervised}). We sketch this method first for completeness.

\subsection{Deep Embedded Clustering}
We choose to apply our constraints formulation to the deep embedded clustering method DEC \cite{xie2016unsupervised} which starts with pre-training an autoencoder ($x_i=g(f(x_i)$) but then removes the decoder. The remaining encoder ($z_i=f(x_i)$) is then fine-tuned by optimizing an objective which takes first $z_i$ and converts it to a soft allocation vector of length $k$ which we term $q_{i,j}$ indicating the degree of belief instance $i$ belongs to cluster $j$. Then $q$ is self-trained on to produce $p$ a unimodal ``hard'' allocation vector which allocates the instance to primarily only one cluster. We now overview each step.

\textbf{Conversion of $z$ to Soft Cluster Allocation Vector $q$.} Here DEC takes the similarity between an embedded point $z_i$ and the cluster centroid $u_j$ measured by Student's $t$-distribution \cite{maaten2008visualizing}. Note that $v$ is a constant as $v = 1$ and $q_{ij}$ is a soft assignment:
\begin{equation}
\label{dec_q}
q_{ij} = \frac{{(1 + {||z_i - {\mu}_{j}||}^{2} / v)} ^ {-\frac{v+1}{2}}} {\sum_{j^{'}} {(1 + {|| z_i - {\mu}_{j^{'}} ||}^{2} / v)} ^ {-\frac{v+1}{2}}}
\end{equation}

\textbf{Conversion of $Q$ To Hard Cluster Assignments $P$.} The above normalized similarities between embedded points and centroids can be considered as soft cluster assignments $Q$. However, we desire a target distribution $P$ that better resembles a hard allocation vector, $p_{ij}$ is defined as:
\begin{equation}
\label{dec_p}
p_{ij} = \frac{ {q_{ij}}^{2} / \sum_{i} q_{ij} } { \sum_{j^{'}} ({q_{ij^{'}}}^{2} / \sum_{i} q_{ij^{'}})}
\end{equation}

\textbf{Loss Function.} Then the algorithm's loss function is to minimize the distance between $P$ and $Q$ as follows. Note this is a form of self-training as we are trying to teach the network to produce unimodal cluster allocation vectors.

\begin{equation}
\label{dec_obj}
\ell_{C} = KL(P || Q) = \sum_{i} \sum_{j} p_{ij} \log{\frac{p_{ij}}{q_{ij}}}
\end{equation}

The DEC method requires the initial centroids given ($\mu$) to calculate $Q$ are ``representative". The initial centroids are set using k-means clustering. However, there is no guarantee that the clustering results over an auto-encoders embedding yield a good clustering. We believe that constraints can help overcome this issue which we test later.

\subsection{Different Types of Constraints}
To enhance the clustering performance and allow for more types of interactions between human and clustering models we propose four types of guidance which are pairwise constraints, instance difficulty constraints, triplet constraints, and cardinality and give examples of each. As traditional constrained clustering methods put constraints on the final clustering assignments, our proposed approach constrains the $q$ vector which is the soft assignment.  A core challenge when adding constraints is to allow the resultant loss function to be differentiable so we can derive back propagation updates. 
\subsubsection{Pairwise Constraints}
Pairwise constraints (must-link and cannot-link) are well studied \cite{basu2008constrained} and we showed they are capable of defining any ground truth set partitions \cite{davidson2007intractability}. Here we show how these pairwise constraints can be added to a deep learning algorithm.
We encode the loss for must-link constraints set ML as:
\begin{equation}
\label{must_link_loss}
\ell_{ML} = -\sum_{(a,b) \in ML} \log{\sum_{j} q_{aj} * q_{bj}}
\end{equation}
Similarly loss for cannot-link constraints set CL is:
\begin{equation}
\label{cannot_link_loss}
\ell_{CL} = -\sum_{(a,b) \in CL} \log{(1 - \sum_{j} q_{aj} * q_{bj})}
\end{equation}
Intuitively speaking, the must-link loss prefers instances with same soft assignments and the cannot-link loss prefers the opposite cases.

\subsubsection{Instance Difficulty Constraints}
A challenge with self-learning in deep learning is that if the initial centroids are incorrect, the self-training can lead to poor results. Here we use constraints to overcome this by allowing the user to specify which instances are easier to cluster (i.e., they belong strongly to only one cluster) and by ignoring difficult instances  (i.e., those that belong to multiple clusters strongly).

We encode user supervision with an $n \times 1$ constraint vector $M$. Let $M_i \in [-1, 1]$ be an instance difficulty indicator, $M_i > 0$ means the instance $i$ is easy to cluster, $M_i = 0$ means no difficulty information is provided and $M_i < 0$ means instance $i$ is hard to cluster. 
The loss function is formulated as:
\begin{equation}
\label{instance_loss}
\ell_{I}   = \sum_{t \in {\{M_t < 0\}}} M_t \sum_{j} {q_{tj}}^2 - \sum_{s \in {\{M_s > 0\}}} M_s\sum_{j} {q_{sj}}^2
 \end{equation}   
The instance difficulty loss function aims to encourage the easier instances to have sparse clustering assignments but prevents the difficult instances having sparse clustering assignments. The absolute value of $M_i$ indicates the degree of confidence in difficulty estimation. This loss will help the model training process converge faster on easier instances and increase our model's robustness towards difficult instances.
 
\subsubsection{Triplet Constraints}
Although pairwise constraints are capable of defining any ground truth set partitions from labeled data \cite{davidson2007intractability}, in many domains no labeled side information exists or strong pairwise guidance is not available. Thus we seek triplet constraints, which are weaker constraints that indicate the relationship within a triple of instances.
Given an anchor instance $a$, positive instance $p$ and negative instance $n$ we say that instance $a$ is more similar to $p$ than to $n$. The loss function for all triplets $(a, p, n) \in T$ can be represented as:
\begin{equation}
\label{triplet_loss}
\ell_{T} = \sum_{(a, p, n) \in T}\max (d(q_a, q_n) - d(q_a, q_p) + \theta, 0)    
\end{equation}
where $d(q_a, q_b) = \sum_j q_{aj} * q_{bj}$ and $\theta > 0$. The larger value of $d(q_a, q_b)$ represents larger similarity between $a$ and $b$. The variable $\theta$ controls the gap distance between positive and negative instances. $\ell_{T}$ works by pushing the positive instance's assignment closer to anchor's assignment and preventing negative instance's assignment being closer to anchor's assignment. 

\subsubsection{Global Size Constraints}
Experts may more naturally give guidance at a cluster level. Here we explore clustering size constraints, which means each cluster should be approximately the same size. Denote the total number of clusters as $k$, total training instances number as $n$, the global size constraints loss function is:
\begin{equation}
\label{global_loss}
\ell_{G} = \sum_{c \in {\{1, .. k\}}}(\sum_{i = 1}^{n} q_{ic}/n - \frac{1}{k})^2
\end{equation}
Our global constraints loss function works by minimizing the distance between the expected cluster size and the actual cluster size. The actual cluster size is calculated by averaging the soft-assignments. To guarantee the effectiveness of global size constraints, we need to assume that during our mini-batch training the batch size should be large enough to calculate the cluster sizes. A similar loss function can be used (see section \ref{sec:extensions}) to enforce other cardinality constraints on the cluster composition such as upper and lower bounds on the number of people with a certain property. 

\subsection{Preventing Trivial Solution}
\label{sec:reconstruction}
In our framework the proposed must-link constraints we mentioned before can lead to trivial solution that all the instances are mapped to the same cluster. Previous deep clustering method \cite{yang2016towards} have also met this problem. To mitigate this problem, we combine the reconstruction loss with the must-link loss to learn together. Denote the encoding network as $f(x)$ and decoding network as $g(x)$, the reconstruction loss for instance $x_i$ is:
\begin{equation}
\label{reconstruction_loss}
\ell_{R} = \ell (g(f(x_i)), x_i)
\end{equation}
where $\ell$ is the least-square loss: $\ell(x, y) = {||x - y||}^2$. 

\subsection{Extensions to High-level Domain Knowledge-Based Constraints}
\label{sec:extensions}
Although most of our proposed constraints are generated based on instance labels or comparisons. Our framework can be extended to high-level domain knowledge-based constraints with minor modifications. 

\textbf{Cardinality Constraints.} For example, cardinality constraints \cite{Dao2016AFF} allow expressing requirements on the number of instances that satisfy some conditions in each cluster. Assume we have $n$ people and want to split them into $k$ dinner party groups. An example cardinality constraint is to enforce each party should have the same number of males and females. 
We split the $n$ people into two groups as $M$ (males) and $F$ (females) in which $|M| + |F| = n$ and $M \cap N = \emptyset$. Then the cardinality constraints can be formulated as:
\begin{equation}
    \label{cardinality_loss}
    \ell_{Cardinality} = \sum_{c \in {\{1, .. k\}}}(\sum_{i \in M} q_{ic}/n - \sum_{j \in F} q_{jc}/n)^2
\end{equation}

For upper-bound and lower-bound based cardinality constraints \cite{Dao2016AFF}, we use the same setting as previously described, now the constraint changes as for each party group we need the number of males to range from $L$ to $U$. Then we can formulate it as:
\begin{equation}
    \label{cardinality_bound_loss}
    \ell_{CardinalityBound} = \sum_{c \in {\{1, .. k\}}} ({\min(0,\sum_{i \in M} q_{ic} - L)}^2 + {\max(0,\sum_{i \in M} q_{ic} - U)}^2)
\end{equation}

\textbf{Logical Combinations of Constraints.} Apart from cardinality constraints, complex logic constraints can also be used to enhance the expressivity power of representing knowledge. For example, if two instances $x_1$ and $x_2$ are in the same cluster then instances $x_3$ and $x_4$ must be in different clusters. This can be achieved in our framework as we can dynamically add cannot-link constraint $CL(x_3, x_4)$ once we check the soft assignment $q$ of $x_1$ and $x_2$.

Consider a horn form constraint like $r \wedge s \wedge t \rightarrow u$. Denote $r = ML(x_1, x_2)$, $s = ML(x_3, x_4)$, $t = ML(x_5, x_6)$ and $u = CL(x_7, x_8)$. By forward passing the instances within $r, s, t$ to our deep constrained clustering model, we can get the soft assignment values of these instances. By checking the satisfying results based on $r \wedge s \wedge t$, we can decide whether to enforce cannot-link loss $CL(x_7, x_8)$.

\section{Putting It All Together - Efficient Training Strategy}
\label{sec:training}
Our training strategy consists of two training branches and effectively has two ways of creating mini-batches for training. For instance-difficulty or global-size constraints, we treat their loss functions as addictive losses so that no extra branch needs to be created. For pairwise or triplet constraints we build another output branch for them and train the whole network in an alternative way. 

\textbf{Loss Branch for Instance Constraints.} In deep learning it is common to add loss functions defined over the same output units. In the Improved DEC method \cite{guo2017improved} the clustering loss $\ell_{C}$ and reconstruction loss $\ell_{R}$ were added together. To this we add the instance difficulty loss $\ell_{I}$. This effectively adds guidance to speed up training convergence by identifying ``easy" instances and increase the model's robustness by ignoring ``difficult" instances. Similarly we treat the global size constraints loss $\ell_{G}$ as an additional additive loss. All instances whether or not they are part of triplet or pairwise constraints are trained through this branch and the mini-batches are created randomly.

\textbf{Loss Branch For Complex Constraints.}
Our framework uses more complex loss functions as they define constraints on pairs and even triples of instances.
Thus we create another loss branch that contains pairwise loss $\ell_{P}$ or triplet loss $\ell_{T}$ to help the network tune the embedding which satisfy these stronger constraints. For each constraint \emph{type} we create a mini-batch consisting of only those instances having that type of constraint.  For each \emph{example} of a constraint type, we feed the constrained instances through the network, calculate the loss, calculate the change in weights but do not adjust the weights. We sum the weight adjustments for all constraint examples in the mini-batch and then adjust the weights. Hence our method is an example of batch weight updating as is standard in DL for stability reasons. The whole training procedure is summarized in Algorithm \ref{alg:dccf}. 
\begin{algorithm}[h]
   \caption{Deep Constrained Clustering Framework}
   \label{alg:dccf}
\begin{algorithmic}
   \STATE {\bfseries Input:} $X$: data, $m$: maximum epochs , $k$: number of clusters, $N$: total number of batches and $N_C$: total number of constraints batches.
   \STATE {\bfseries Output:} latent embeddings $Z$, cluster assignment $S$.
   \smallskip
  \STATE Train the stacked denosing autoencoder to obtain $Z$
  \STATE Initialize centroids $\mu$ via k-means on embedding $Z$.
   \FOR{$epoch=1$ {\bfseries to} $m$}
    \FOR{$batch=1$ {\bfseries to} $N$}
   \STATE Calculate $\ell_{C}$ via Eqn (\ref{dec_obj}), $\ell_{R}$ via Eqn (\ref{reconstruction_loss}). 
   \STATE Calculate $\ell_{I}$ via Eqn (\ref{instance_loss}) or $\ell_{G}$ via Eqn (\ref{global_loss}).
   \STATE Calculate total loss as $\ell_{C} + \ell_{R} + \{ \ell_{I} || \ell_{G}\}$.
   \STATE Update network parameters based on total loss. 
   \ENDFOR
   \FOR{$batch=1$ {\bfseries to} $N_C$}
   \STATE Calculate $\ell_{P}$ via Eqn (\ref{must_link_loss}, \ref{cannot_link_loss}) or $\ell_{T}$ via Eqn (\ref{triplet_loss}). 
   \STATE Update network parameters based on $\{\ell_{P} || \ell_{T}\}$ . 
   \ENDFOR
   \STATE Forward pass to compute $Z$ and $S_i = \argmax_{j} q_{ij}$.
   \ENDFOR
\end{algorithmic}
\end{algorithm}

\section{Experiments}
\label{sec:experiments}
All data and code used to perform these experiments are available online (\url{http://github.com/blueocean92/deep_constrained_clustering}) to help with reproducibility. In our experiments we aim to address the following questions:
\begin{itemize}
    \item How does our end-to-end deep clustering approach using traditional pairwise constraints compare with traditional constrained clustering methods? The latter is given the same auto-encoding representation $Z$ used to initialize our method. 
    \item Are the new types of constraints we create for deep clustering method useful in practice?
    \item Is our end-to-end deep constrained clustering method more robust to the well known negative effects of constraints we published earlier \cite{davidson2006measuring}?
\end{itemize}

    \subsection{Datasets} 
        To study the performance and generality of different algorithms, we evaluate the proposed method on two image datasets and one text dataset:

\noindent
\textbf{MNIST}: Consists of $70000$ handwritten digits of $28$-by-$28$ pixel size. The digits are centered and size-normalized in our experiments \cite{lecun1998gradient}. 

\noindent
\textbf{FASHION-MNIST}: A Zalando's article images-consisting of a training set of $60000$ examples and a test set of $10000$ examples. Each example is a $28$-by-$28$ grayscale image, associated with a label from $10$ classes. 

\noindent
\textbf{REUTERS-10K}:  This dataset contains English news stories labeled with a category tree \cite{lewis2004rcv1}. To be comparable with the previous baselines, we used $4$ root categories: \texttt{corporate/industrial, government/social, markets} and \texttt{economics} as labels and excluded all documents with multiple labels. We randomly sampled a subset of $10000$ examples and computed TF-IDF features on the $2000$ most common words. 

\subsection{Implementation Details}
\textbf{Basic Deep Clustering Implementation.} To be comparable with deep clustering baselines, we set the encoder network as a fully connected multilayer perceptron with dimensions $d-500-500-2000-10$ for all datasets, where $d$ is the dimension of input data(features). The decoder network is a mirror of the encoder. All the internal layers are activated by the ReLU \cite{nair2010rectified} nonlinearity function. For a fair comparison with baseline methods, we used the same greedy layer-wise pre-training strategy to calculate the auto-encoders embedding. To initialize clustering centroids, we run k-means with 20 restarts and select the best solution. We choose Adam optimizer with an initial learning rate of $0.001$ for all the experiments. We adopt standard metrics for evaluating clustering performance which measure how close the clustering found is to the ground truth result. Specifically, we employ the following two metrics: normalized mutual information(\textbf{NMI})\cite{strehl2000impact,xu2003document} and clustering accuracy(\textbf{Acc})\cite{xu2003document}. In our baseline comparisons we use IDEC \cite{guo2017improved}, a non-constrained improved version of DEC published recently.

\textbf{Pairwise Constraints Experiments.} We randomly select pairs of instances and generate the corresponding pairwise constraints between them. 
To ensure transitivity we calculate the transitive closure over all must-linked instances and then generate entailed constraints from the cannot-link constraints \cite{davidson2007intractability}. Since our loss function for must-link constraints is combined with reconstruction loss, we use grid search and set the penalty weight for must-link as $0.1$. 
        
\begin{figure}[h]
        \centering
        \subfigure{\includegraphics[width=0.6\columnwidth]{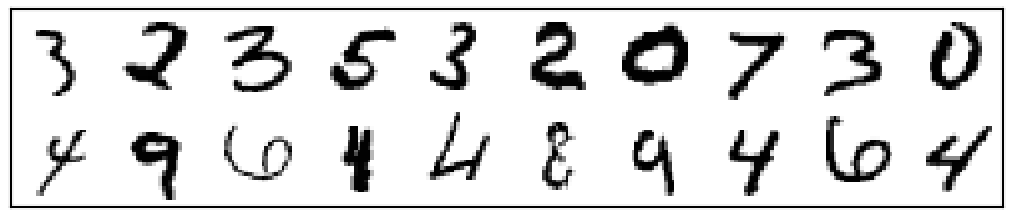}}
        \caption{Example of instance difficulty constraints. Top row shows the ``easy'' instances and second row shows the ``difficult'' instances. }
        \label{fig:instance_mnist}
    \end{figure}

\textbf{Instance Difficulty Constraints Experiments.} To simulate human-guided instance difficulty constraints, we use k-means as a base learner and mark all the incorrectly clustered instances as difficult with confidence $0.1$, we also mark the correctly classified instances as easy instances with confidence $1$. 
In Figure \ref{fig:instance_mnist} we give some example difficulty constraints found using this method. 
    
\begin{figure}[h]
        \centering
              \subfigure{\includegraphics[width=0.48\columnwidth]{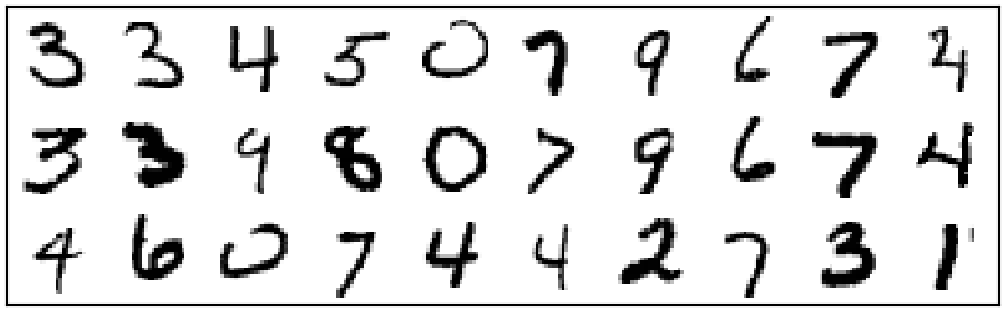}}
            \hfill
              \subfigure{\includegraphics[width=0.48\columnwidth]{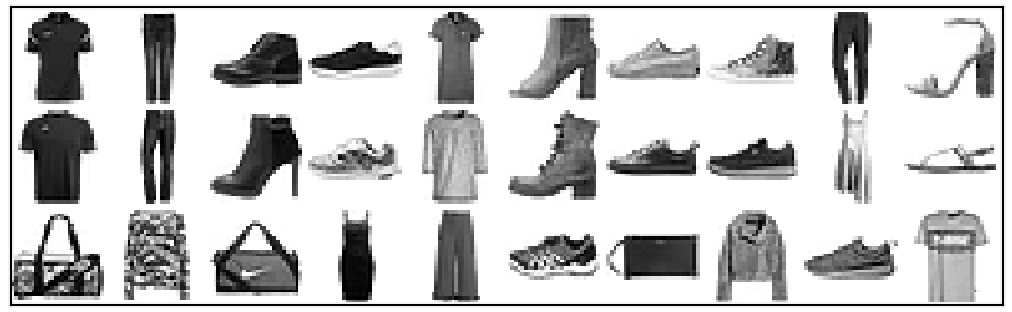}}
        \caption{Examples of the generated triplet constraints for MNIST and Fashion. The three rows for each plot shows the anchor instances, positive instances and negative instances correspondingly.}
               \label{fig:triplet_visual}
        \end{figure}

\textbf{Triplet Constraints Experiments.} Triplet constraints can state that instance $i$ is more similar to instance $j$ than instance $k$. To simulate human guidance on triplet constraints, we randomly select $n$ instances as anchors ($i$), for each anchor we randomly select two instances ($j$ and $k$) based on the similarity between the anchor. The similarity is calculated as the euclidian distance $d$ between two instances pre-trained embedding. The pre-trained embedding is extracted from our deep clustering network trained with $100000$ pairwise constraints. 
Figure \ref{fig:triplet_visual} shows the generated triplets constraints. Through grid search we set the triplet loss margin $\theta = 0.1$.
        
\textbf{Global Size Constraints Experiments.} We apply global size constraints to MNIST and Fashion datasets since they satisfy the balanced size assumptions. The total number of clusters is set to $10$. 

\subsection{Experimental Results}
\label{sec:exp_results}
\textbf{Experiments on Instance Difficulty.}
        In Table \ref{tab:instance}, we report the average test performance of deep clustering framework without any constraints in the left. In comparison, we report the average test performance of deep clustering framework with instance difficulty constraints in the right and we find the model learned with instance difficulty constraints outperforms the baseline method in all datasets. This is to be expected as we have given the algorithm more information than the baseline method, but it demonstrates our method can make good use of this extra information. What is unexpected is the effectiveness of speeding up the learning process and will be the focus of future work.           
        \begin{table}[htb]
        \begin{minipage}{.48\linewidth}
        \centering
            \resizebox{\columnwidth}{!}{%
            \begin{tabular}{cccc}
            \toprule
                 & MNIST & Fashion & Reuters\\
                \midrule
                Acc(\%)          &    $88.29 \pm 0.05$    &$58.74 \pm 0.08$     &$75.20 \pm 0.07$\\
                NMI(\%)        &    $86.12 \pm 0.09$    &$63.27 \pm 0.11$     &$54.16 \pm 1.73$ \\
                Epoch      &    $87.60 \pm 12.53$         & $77.20 \pm 11.28$        &$12.90 \pm 2.03$\\
                \bottomrule
            \end{tabular}
            }
            
        \end{minipage}
        \hfill
        \begin{minipage}{.48\linewidth}
        \centering
            \resizebox{\columnwidth}{!}{%
            \begin{tabular}{cccc}
            \toprule
                 & MNIST & Fashion & Reuters\\
                \midrule
                Acc(\%)          &    $91.02 \pm 0.34$    &$62.17 \pm 0.06$     &$78.01 \pm 0.13$\\
                NMI(\%)        &    $88.08 \pm 0.14$    &$64.95 \pm 0.04$     &$56.02 \pm 0.21$ \\
                Epoch      &    $29.70 \pm 4.25$         & $47.60 \pm 6.98$        &$9.50 \pm 1.80$\\
                \bottomrule
            \end{tabular}
            }
        \end{minipage} 
        \caption{Left table shows baseline results for Improved DEC \cite{guo2017improved} averaged over $20$ trials. Right table lists experiments using instance difficulty constraints (mean $\pm$ std) averaged over $20$ trials. }
        \label{tab:instance}
        \end{table}
        
    \begin{figure*}[ht]
        \centering
              \subfigure[MNIST]{\includegraphics[width=0.32\textwidth]{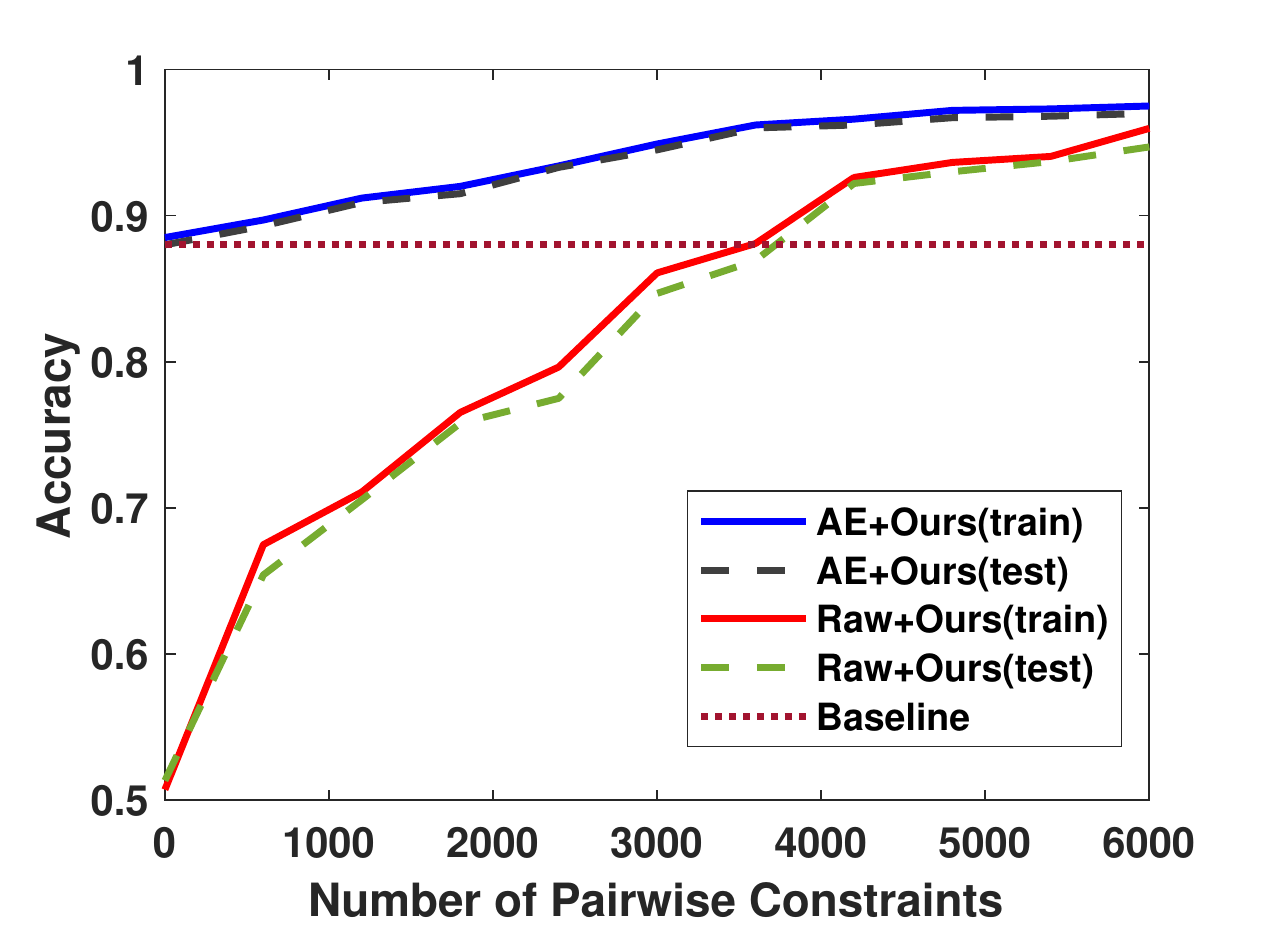}}
            \hfill
              \subfigure[Fashion]{\includegraphics[width=0.32\textwidth]{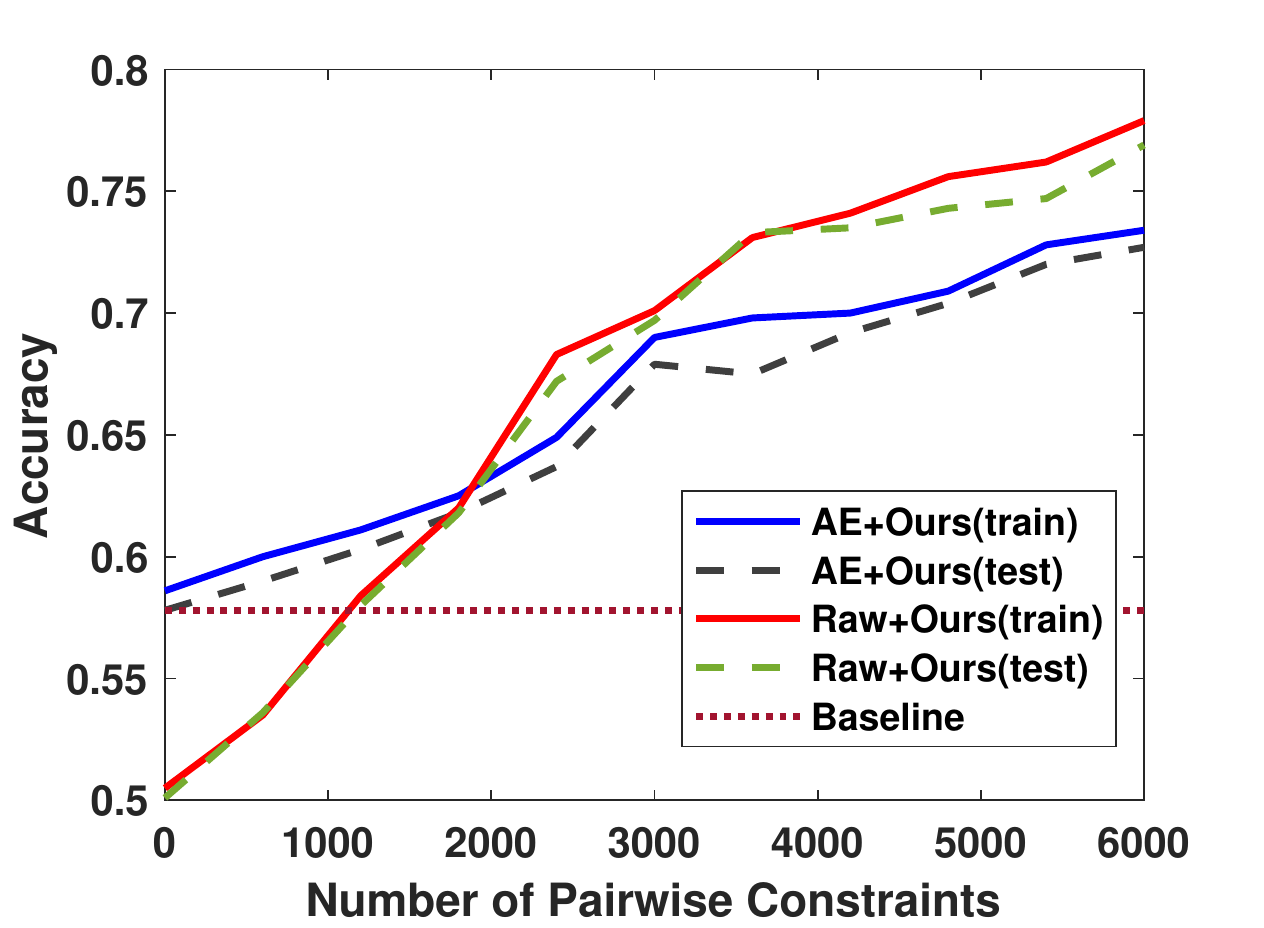}}
            \hfill
             \subfigure[Reuters]{\includegraphics[width=0.32\textwidth]{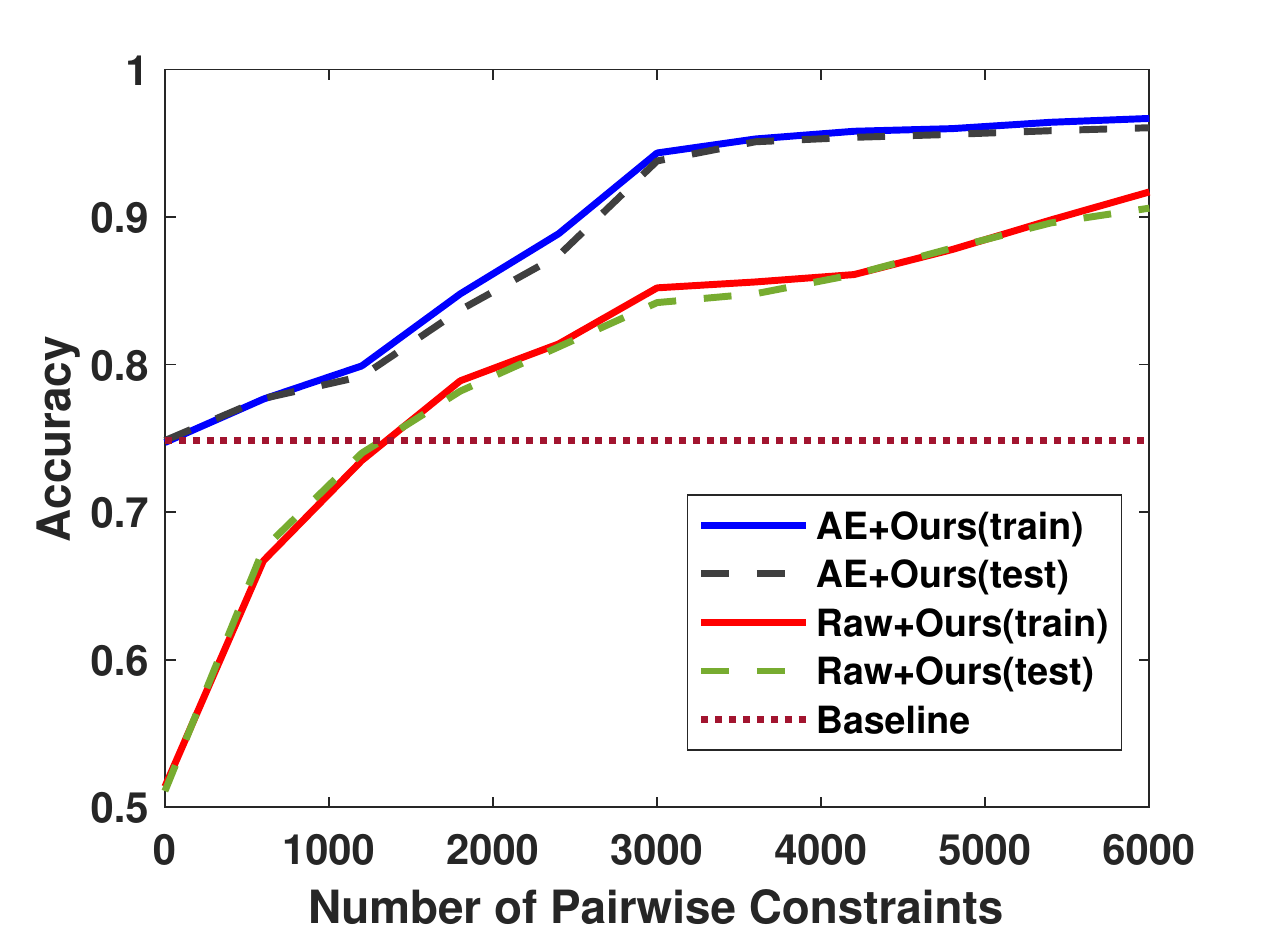}}
        
        \subfigure[MNIST]{\includegraphics[width=0.32\textwidth]{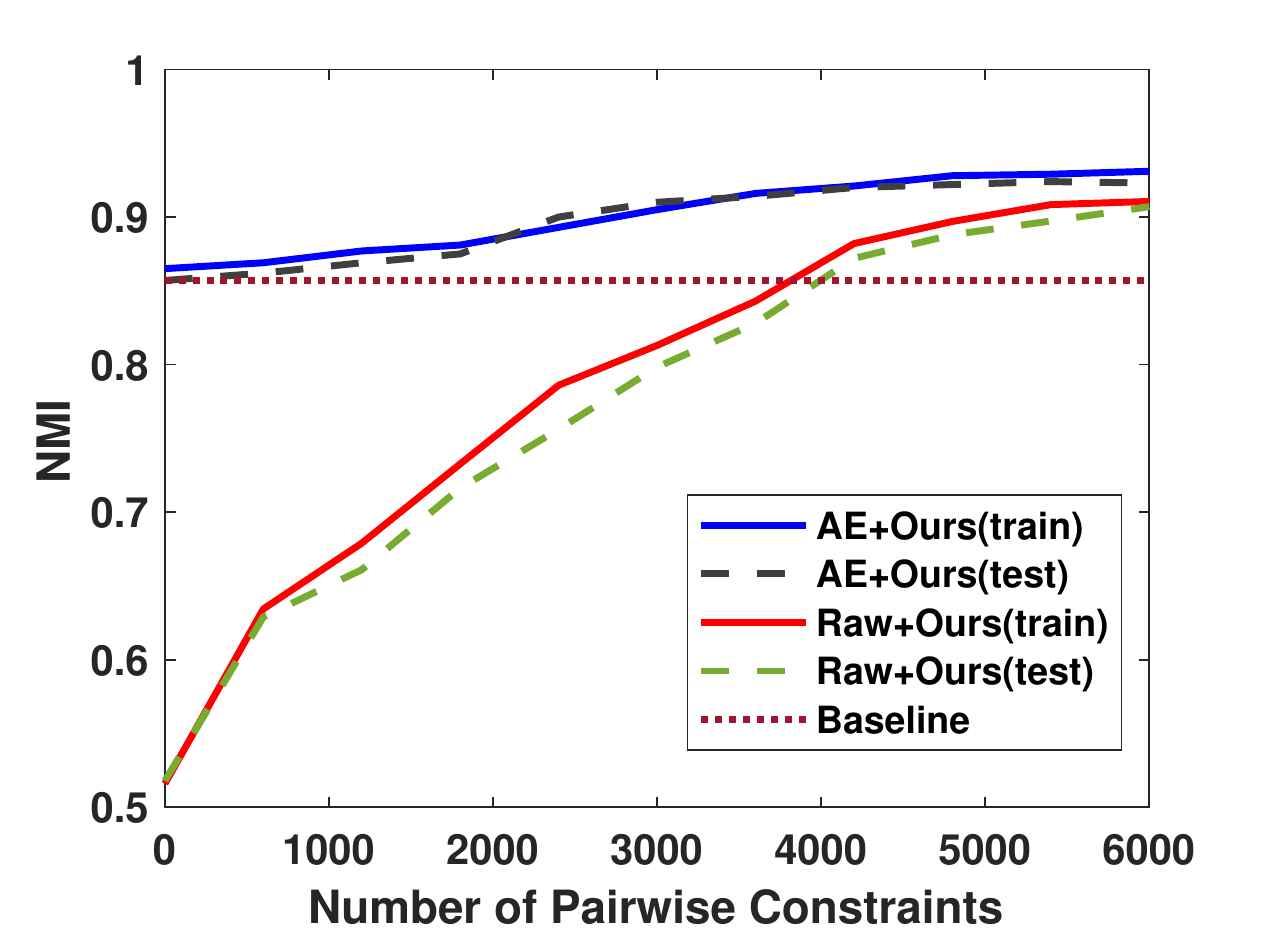}}
            \hfill
              \subfigure[Fashion]{\includegraphics[width=0.32\textwidth]{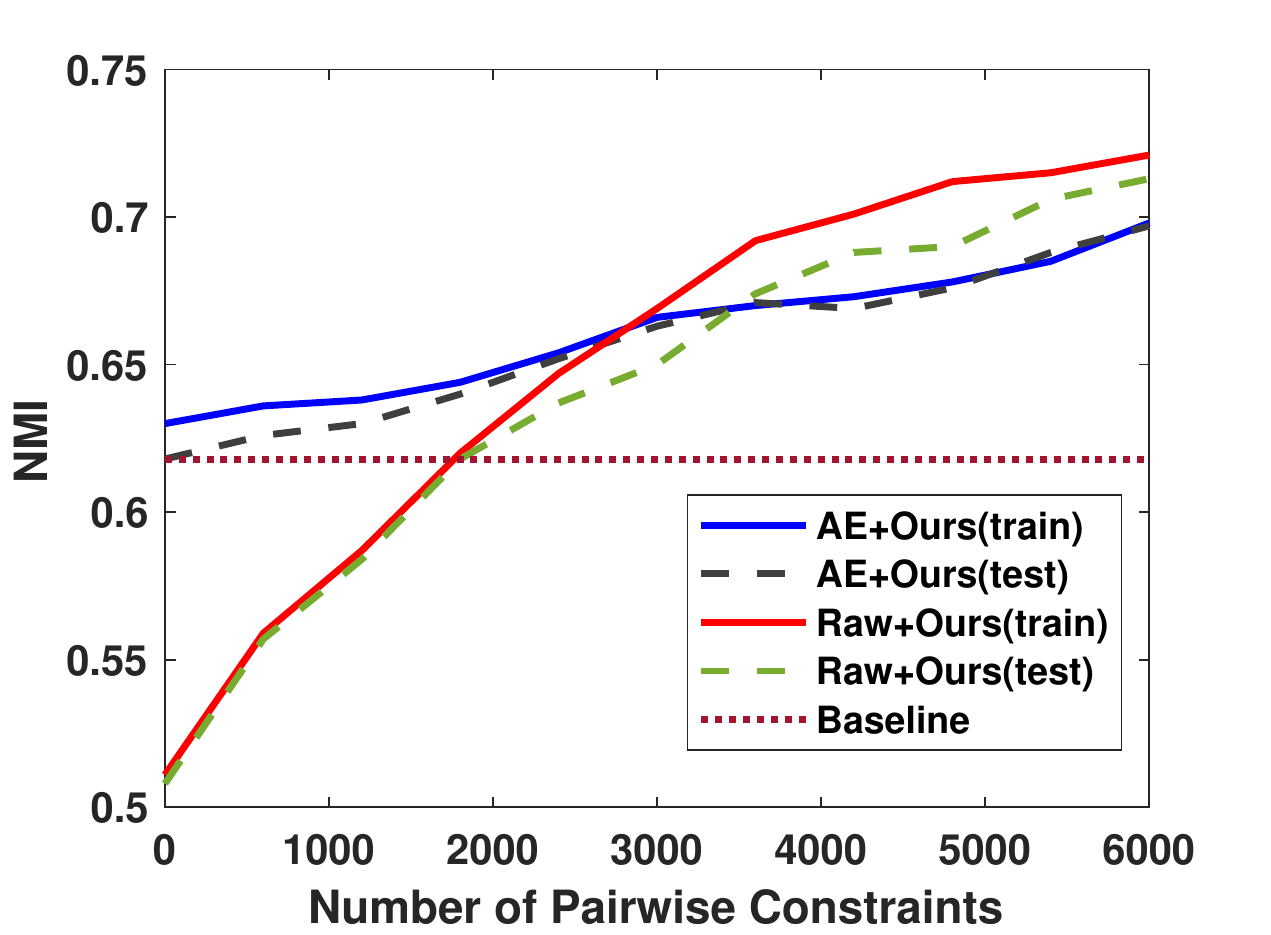}}
            \hfill
             \subfigure[Reuters]{\includegraphics[width=0.32\textwidth]{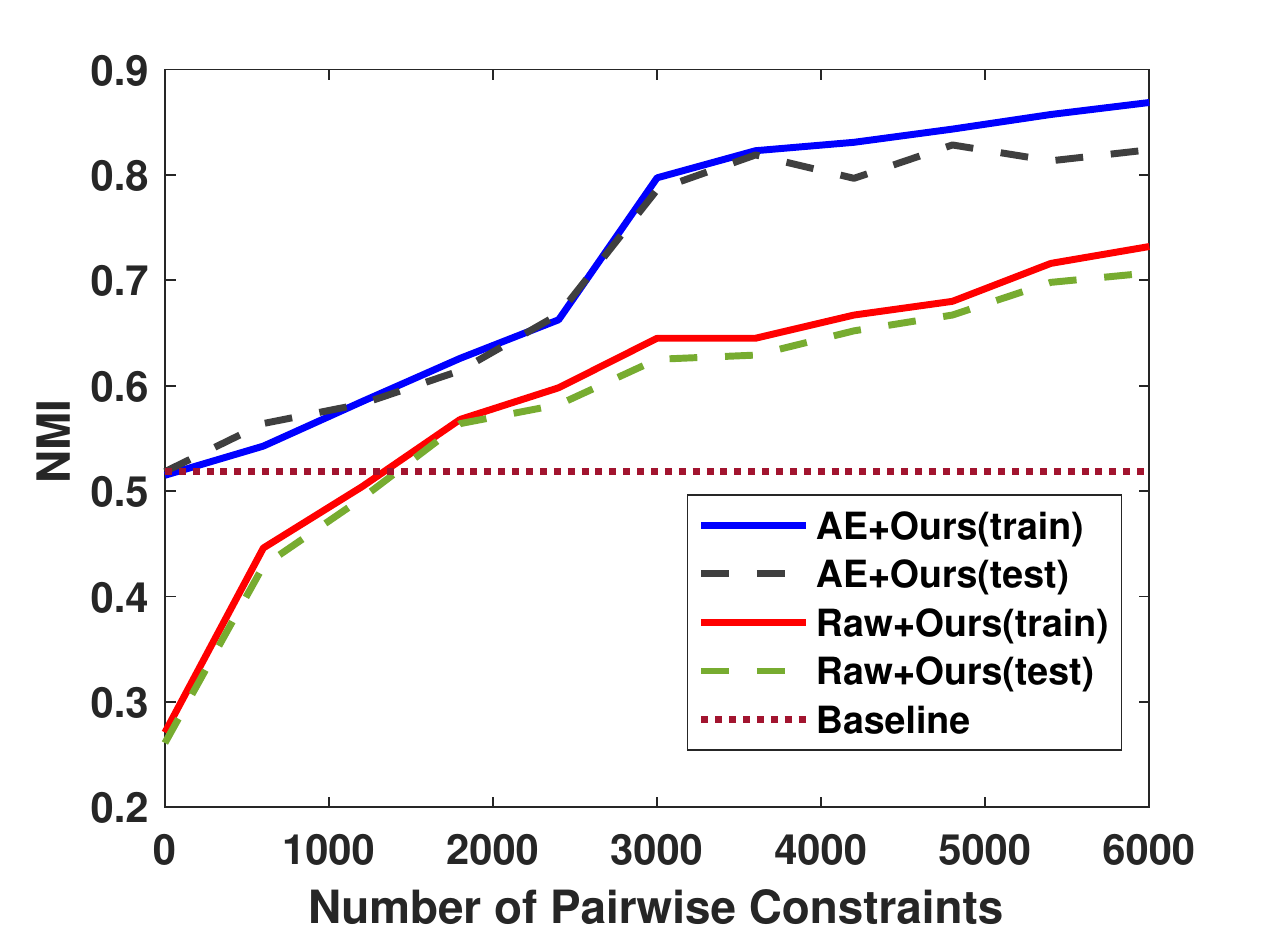}}
        \caption{Clustering accuracy and NMI on training test sets for different number of pairwise constraints. AE means an autoencoder was used to seed our method. The horizontal maroon colored baseline shows the IDEC's \cite{guo2017improved} test set performance.}
        \label{fig:pairwise_performance}
    \end{figure*}
\subsubsection{Experiments on pairwise constraints}
 We randomly generate $6000$ pairs of constraints which are a small fractions of possible pairwise constraints for MNIST ($0.0002 \%$), Fashion ($0.0002 \%$) and Reuters ($0.006 \%$).

Recall the DEC method is initialized with auto-encoder features. To better understand the contribution of pairwise constraints, we have tested our method with both auto-encoders features and raw data. As can be seen from Figure \ref{fig:pairwise_performance}: the clustering performance improves consistently as the number of constraints increases in both settings. Moreover, with just $6000$ pairwise constraints the performance on Reuters and MNIST increased significantly especially for the setup with raw data. We also notice that learning with raw data in Fashion achieves a better result than using autoencoder's features. This shows that the autoencoder's features may not always be suitable for DEC's clustering objective. Overall our results show pairwise constraints can help reshape the representation and improve the clustering results. 

We also compare the results with recent work \cite{hsu2015neural}: our approach(autoencoders features) outperforms the best clustering accuracy reported for MNIST by a margin of $16.08\%$, $2.16\%$ and $0.13\%$ respectively for 6, 60 and 600 samples/class. Unfortunately, we can't make a comparison with Fogel's algorithm \cite{fogel2018clustering} due to an issue in their code repository.
    \begin{table}[H]
        \begin{center}
        \begin{tabular}{ccccc}
            \toprule[1.6pt]
                 &Flexible CSP{*}& COP-KMeans & MPCKMeans & Ours \\
                \midrule
                MNIST Acc            &    $0.628 \pm 0.07$&    $0.816 \pm 0.06$    & $0.846 \pm 0.04$    &{$\textbf{0.963} \pm \textbf{0.01}$} \\
                MNIST NMI            &    $0.587 \pm 0.06$&    $0.773 \pm 0.02$    & $0.808 \pm 0.04$    &{$\textbf{0.918} \pm \textbf{0.01}$} \\
                Negative Ratio     &    $19 \%$&        $45 \%$            & $11 \%$                &{$\textbf{0 \%}$}\\
                \midrule
                Fashion Acc            &$0.417 \pm 0.05$  &    $0.548 \pm 0.04$    & $0.589 \pm 0.05$    &{$\textbf{0.681} \pm \textbf{0.03}$}\\
                Fashion NMI           &$0.462 \pm 0.03$ &    $0.589 \pm 0.02$    & $0.613 \pm 0.04$    &{$\textbf{0.667} \pm \textbf{0.02}$}\\
                Negative Ratio     & $23 \%$&        $27 \%$            & $37 \%$                &{$\textbf{6 \%}$}\\
                \midrule
                Reuters Acc            &$0.554 \pm 0.07$ &    $0.712 \pm 0.0424$    & $0.763 \pm 0.05$    &{$\textbf{0.950} \pm \textbf{0.02}$}\\
                Reuters NMI            &$0.410 \pm 0.05$ &    $0.478 \pm 0.0346$    & $0.544 \pm 0.04$    &{$\textbf{0.815} \pm \textbf{0.02}$}\\
                Negative Ratio     &$28 \%$ &        $73 \%$            & $80 \%$                &{$\textbf{0 \%}$}\\
                \bottomrule[1.6pt]
            \end{tabular}
            \end{center}
            \caption{Pairwise constrained clustering performance (mean $\pm$ std) averaged over $100$ constraints sets. Due to the scalability issues we apply flexible CSP with downsampled data($3000$ instances and $180$ constraints). Negative ratio is the fraction of times using constraints resulted in poorer results than not using constraints. See Figure \ref{fig:visual_embedding} and text for an explanation why our method performs well.}
            \label{pairwise_neg}
            \end{table}
        \begin{figure}[h]
        \centering
              \subfigure[MNIST (AE)]{\includegraphics[width=0.325\columnwidth]{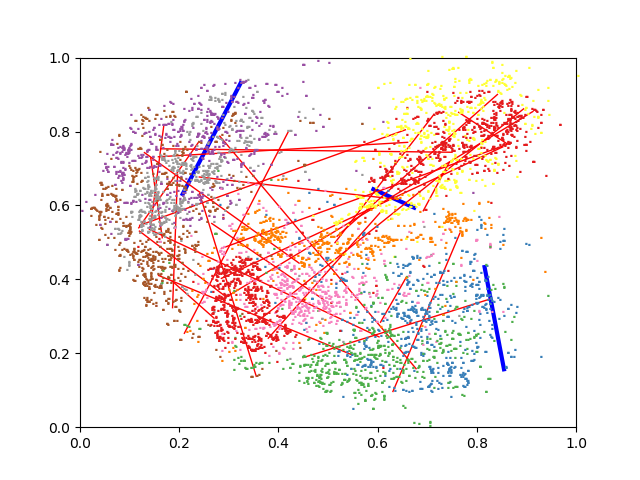}}
            \hfill
              \subfigure[MNIST (IDEC)]{\includegraphics[width=0.325\columnwidth]{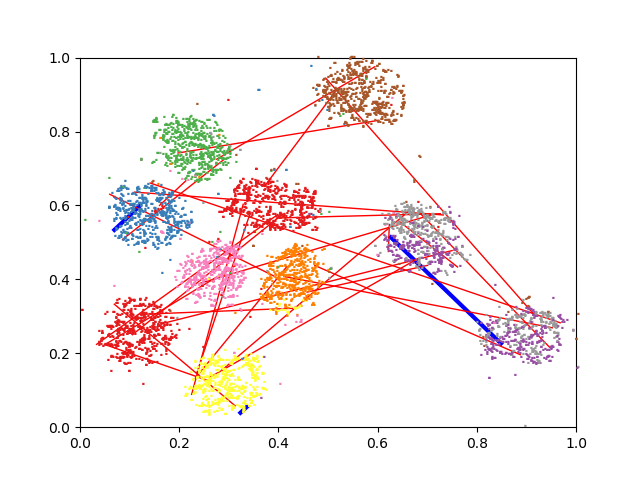}}
            \hfill
             \subfigure[MNIST (Ours)]{\includegraphics[width=0.325\columnwidth]{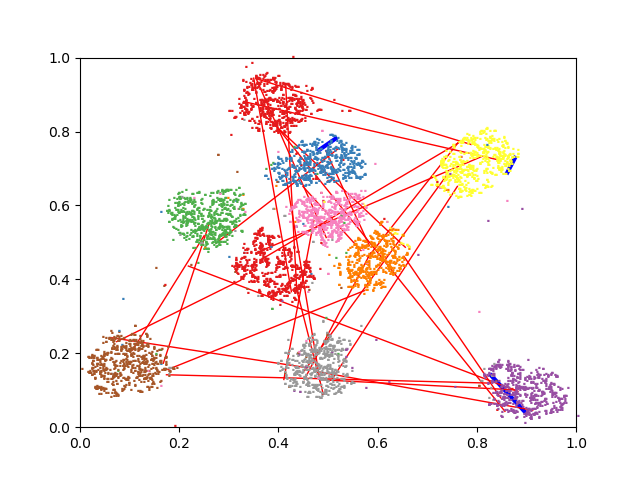}}
        
             \subfigure[Fashion (AE)]{\includegraphics[width=0.325\columnwidth]{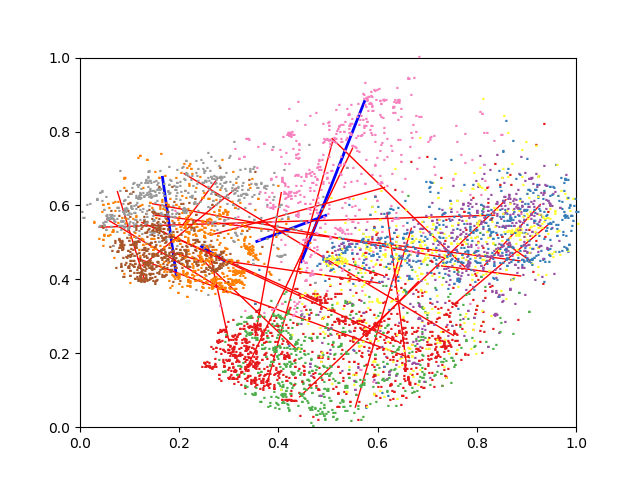}}
            \hfill
              \subfigure[Fashion (IDEC)]{\includegraphics[width=0.33\columnwidth]{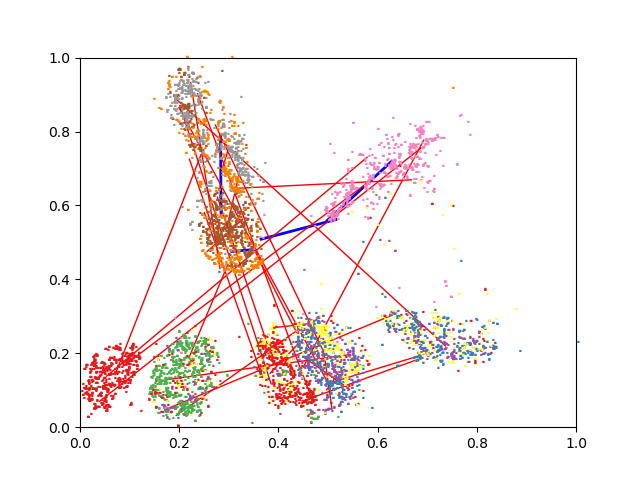}}
            \hfill
             \subfigure[Fashion (Ours)]{\includegraphics[width=0.325\columnwidth]{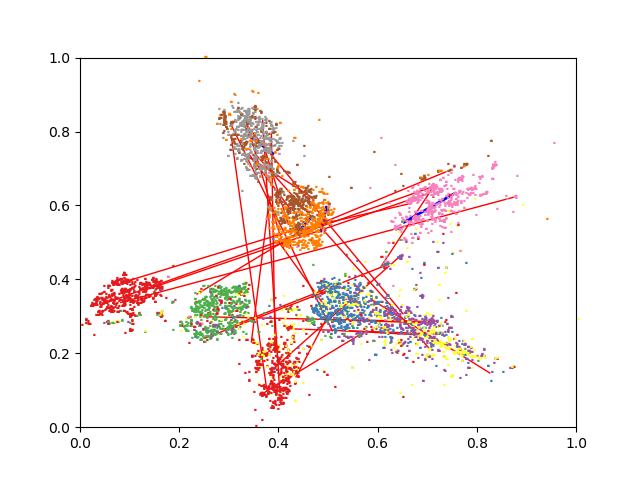}}    
        
            \subfigure[Reuters (AE)]{\includegraphics[width=0.325\columnwidth]{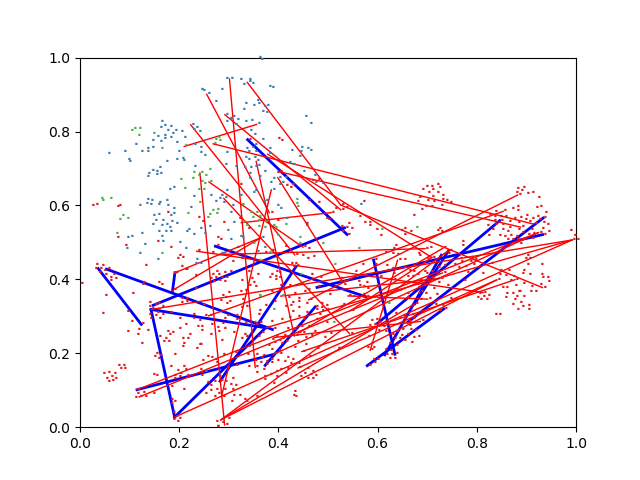}}
            \hfill
            \subfigure[Reuters (IDEC)]{\includegraphics[width=0.325\columnwidth]{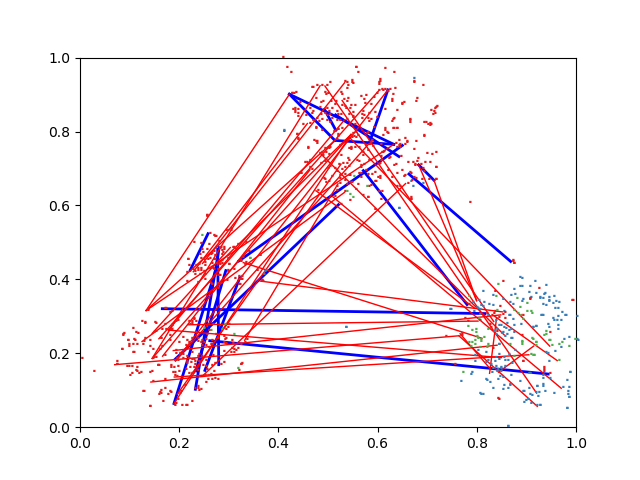}}
            \hfill
            \subfigure[Reuters (Ours)]{\includegraphics[width=0.325\columnwidth]{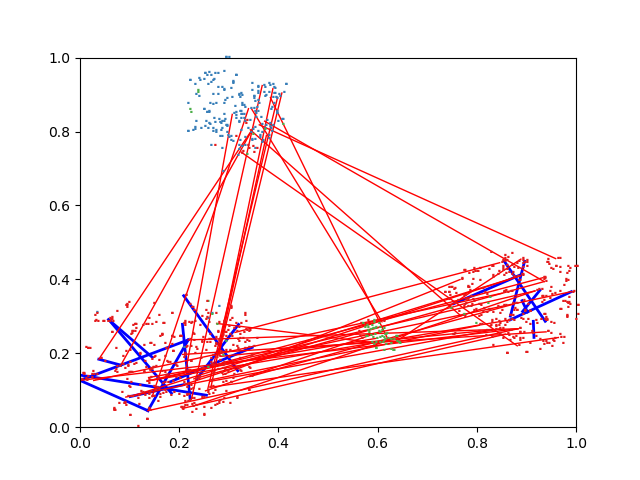}}
        \caption{We visualize (using t-SNE) the latent representation for a subset of instances and pairwise constraints, we visualize the same instances and constraints for each row. The red lines are cannot-links and blue lines are must-links.}
        \label{fig:visual_embedding}
        \end{figure}
\textbf{Negative Effects of Constraints.} Our earlier work \cite{davidson2006measuring} showed that for traditional constrained clustering algorithms, that the addition of constraints \emph{on average} helps clustering but many individual constraint sets can hurt performance in that  performance is worse than using \textbf{no} constraints. Here we recreate these results even when these classic methods use auto-encoded representations. In Table \ref{pairwise_neg}, we report the average performance with $3600$ randomly generated pairwise constraints. For each dataset, we randomly generated $100$ sets of constraints to test the negative effects of constraints\cite{davidson2006measuring}. In each run, we fixed the random seed and the initial centroids for k-means based methods, for each method we compare its performance between constrained version to unconstrained version. We calculate the negative ratio which is the fraction of times that unconstrained version produced better results than the constrained version. As can be seen from the table, our proposed method achieves significant improvements than traditional non-deep constrained clustering algorithms \cite{wagstaff2001constrained,bilenko2004integrating,wang2010flexible}.        

To understand why our method was robust to variations in constraint sets we visualized the embeddings learnt.        Figure \ref{fig:visual_embedding} shows the embedded representation of a random subset of instances and its corresponding pairwise constraints using t-SNE and the learned embedding $z$. Based on Figure \ref{fig:visual_embedding}, we can see the autoencoders embedding is noisy and lot's of constraints are inconsistent based on our earlier definition  \cite{davidson2006measuring}. Further, we visualize the IDEC's latent embedding and find out the clusters are better separated. However, the inconsistent constraints still exist (blue lines across different clusters and redlines within a cluster); these constraints tend to have negative effects on traditional constrained clustering methods. Finally, for our method's results we can see the clusters are well separated, the must-links are well satisfied(blue lines are within the same cluster) and cannot-links are well satisfied(red lines are across different clusters).  Hence we can conclude that end-to-end-learning can address these negative effects of constraints by simultaneously learning a representation that is consistent with the constraints and clustering the data.       
This result has profound practical significance as practitioners typically only have one constraint set to work with.
        \subsubsection{Experiments on triplet constraints}
\begin{figure}[h]
        \centering
              \subfigure[MNIST(Triplet)]{\includegraphics[width=0.45\columnwidth]{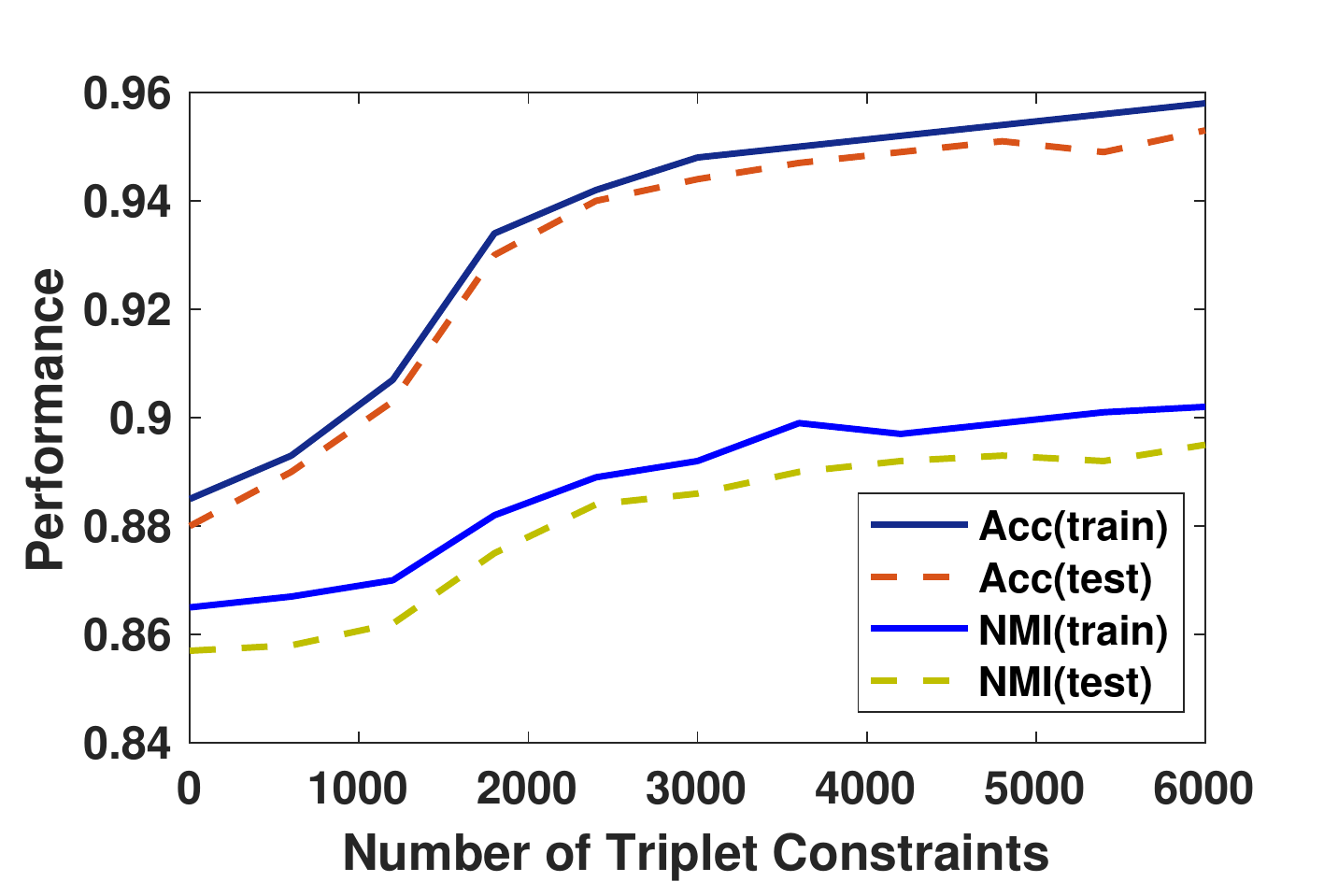}}
            \hfill
              \subfigure[Fashion(Triplet)]{\includegraphics[width=0.45\columnwidth]{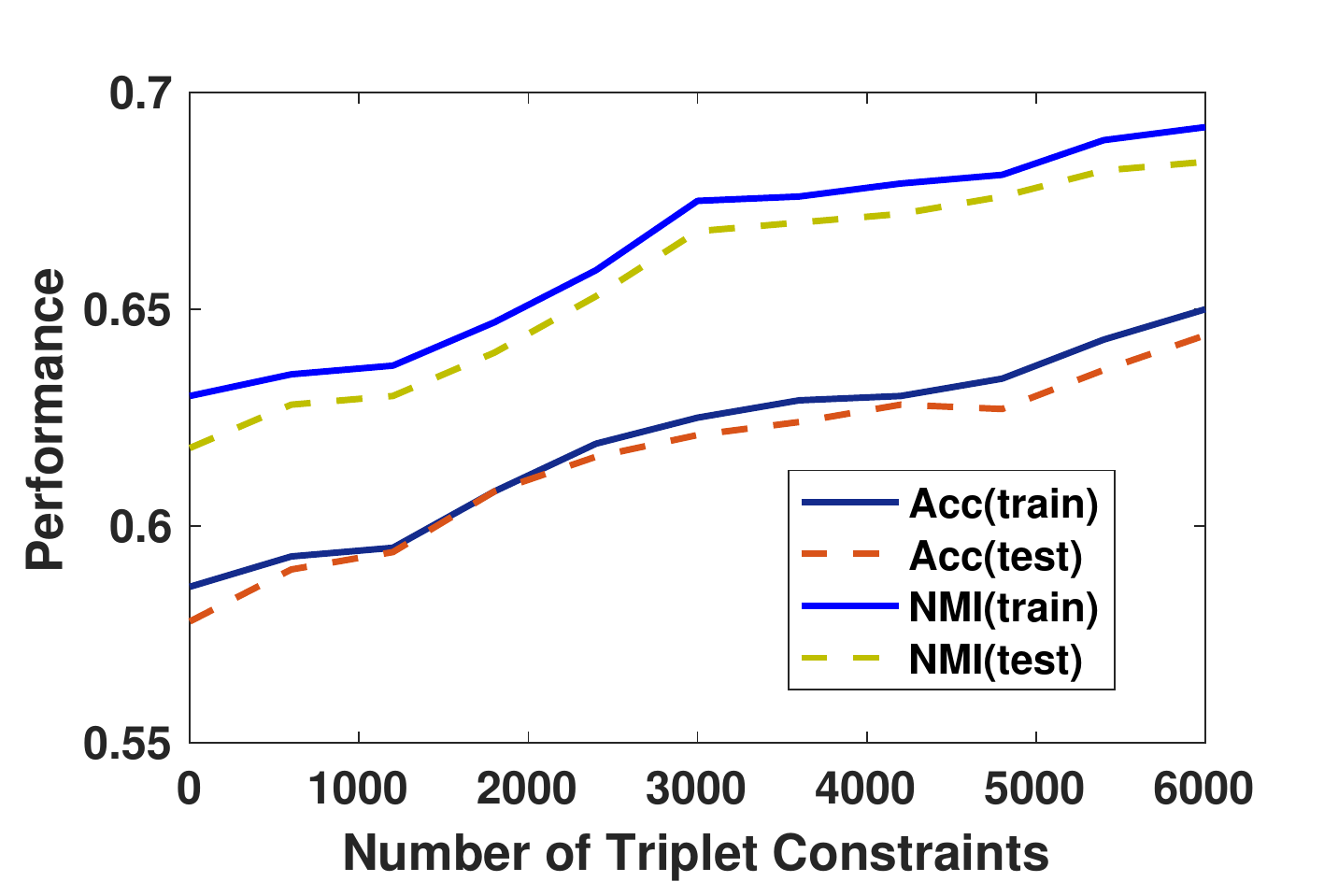}}
        \caption{Evaluation of the effectiveness of triplet constraints in terms of Acc/NMI. }
                \label{fig:triplet_performance}
        \end{figure}

        We experimented on MNIST and FASHION datasets. Figure \ref{fig:triplet_visual} visualizes example triplet constraints (based on embedding similarity), note the positive instances are closer to anchors than negative instances. In Figure \ref{fig:triplet_performance}, we show the clustering Acc/NMI improves consistently as the number of constraints increasing. Comparing with Figure \ref{fig:pairwise_performance} we can find the pairwise constraints can bring slightly better improvements, that's because our triplets constraints are generated from a continuous domain and there is no exact together/apart information encoded in the constraints. Triplet constraints can be seen as a weaker but more general type of constraints.
        
        \subsubsection{Experiments on global size constraints}
\begin{figure}[ht]
        \centering
              \subfigure[MNIST]{\includegraphics[width=0.45\columnwidth]{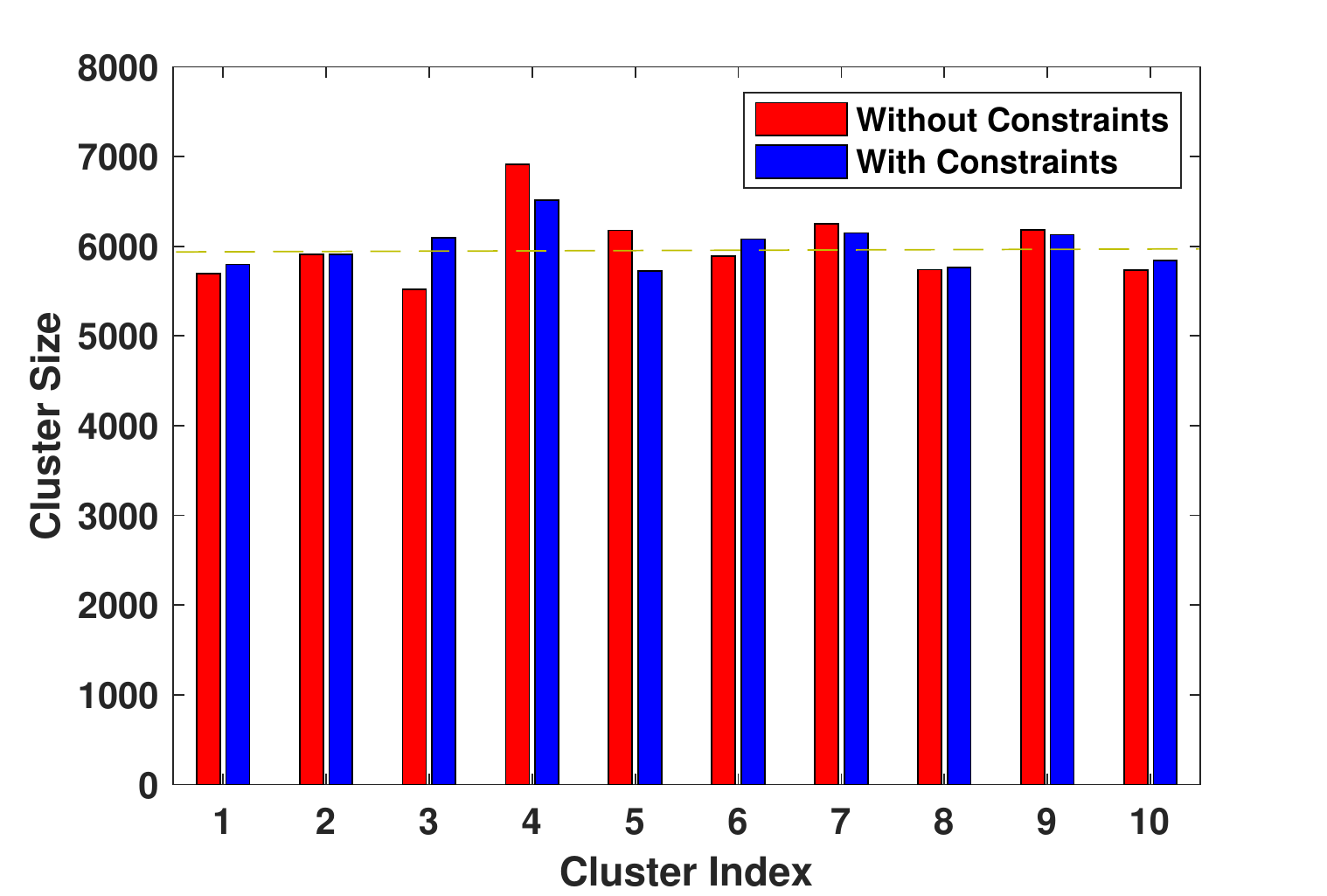}}
            \hfill
              \subfigure[Fashion]{\includegraphics[width=0.45\columnwidth]{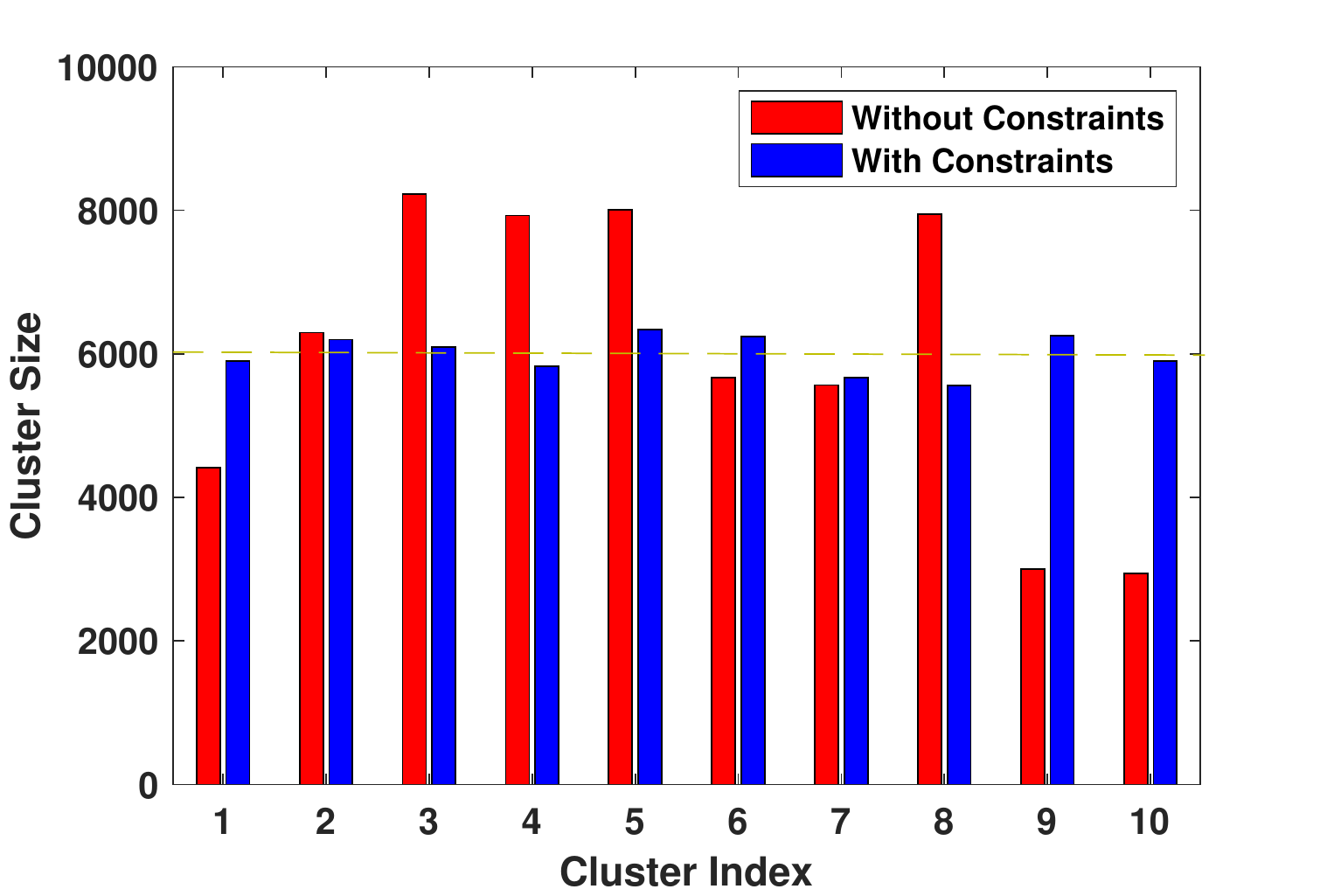}}
        \caption{Evaluation of the global size constraints. This plot shows each cluster's size before/after adding global size constraints.}
        \label{fig:global_constraints}
        \end{figure}

        To test the effectiveness of our proposed global size constraints, we have experimented on MNIST and Fashion training set since they both have balanced cluster sizes (see Figure \ref{fig:global_constraints}). Note that the ideal size for each cluster is $6000$ (each data set has $10$ classes), we can see that blue bars are more evenly distributed and closer to the ideal size.
        
        We also evaluate the clustering performance with global constraints on MNIST (Acc:$0.91$, NMI:$0.86$) and Fashion (Acc:$0.57$, NMI:$0.59$). Comparing to the baselines in table \ref{tab:instance}, interestingly, we find the performance improved slightly on MNIST but dropped slightly on Fashion. 

\section{Conclusion and Future Work}
\label{sec:conclusion}
The area of constrained partitional clustering has a long history and is widely used. Constrained partitional clustering typically is mostly limited to simple pairwise together and apart constraints. In this paper, we show that deep clustering can be extended to a variety of fundamentally different constraint types including instance-level (specifying hardness), cluster level (specifying cluster sizes) and triplet-level. 

Our deep learning formulation was shown to advance the general field of constrained clustering in several ways. Firstly, it achieves better experimental performance than well-known k-means, mixture-model and spectral constrained clustering in both an academic setting and a practical setting (see Table \ref{pairwise_neg}). 

Importantly, our approach does not suffer from the negative effects of constraints \cite{davidson2006measuring} as it learns a representation that simultaneously satisfies the constraints and finds a good clustering. This result is quite useful as a practitioner typically has just one constraint set and our method is far more likely to perform better than using no constraints. 

Most importantly, we were able to show that our method achieves all of the above but still retains the benefits of deep learning such as scalability, out-of-sample predictions and end-to-end learning. We found that even though standard non-deep learning methods were given the same representations of the data used to initialize our methods the deep constrained clustering was able to adapt these representations even further.
Future work will explore new types of constraints, using multiple constraints at once and extensions to the clustering setting.

\section*{Acknowledgements}
We acknowledge support for this work from a Google Gift entitled: ``Combining Symbolic Reasoning and Deep Learning".

\end{document}